\documentclass[lettersize,journal]{IEEEtran}
\usepackage{amsmath,amsfonts}
\usepackage{algorithmic}
\usepackage{algorithm}
\usepackage{array}
\usepackage[caption=false,font=normalsize,labelfont=sf,textfont=sf]{subfig}
\usepackage{textcomp}
\usepackage{stfloats}
\usepackage{url}
\usepackage{verbatim}
\usepackage{graphicx}
\usepackage{cite}
\usepackage{mathptmx} % assumes new font selection scheme installed
\usepackage{times} % assumes new font selection scheme installed
\usepackage{amssymb}  % assumes amsmath package installed
\usepackage{color}
\usepackage[dvipsnames]{xcolor}
\usepackage{romannum}
\usepackage{comment}
\usepackage{physics}

\begin{document}

\title{Breaking the Circle: An Autonomous Control-Switching Strategy for Stable Orographic Soaring in MAVs}

\author{Sunyou Hwang$^{1}$, Christophe De Wagter$^{1}$, Bart Remes$^{1}$, and Guido de Croon$^{1}$
        % <-this % stops a space
\thanks{$^{1}$ All authors are with the MAVLab, Department of Control and Operations, Faculty of Aerospace Engineering, Delft University of Technology, 2629 HS Delft, the Netherlands.}% <-this % stops a space
% {\tt\footnotesize (email: S.Hwang-1@tudelft.nl, C.dewagter@tudelft.nl, B.D.W.Remes@tudelft.nl, G.C.H.E.deCroon@tudelft.nl)}
%\thanks{Manuscript received April 19, 2021; revised August 16, 2021.}
}

\markboth{Journal of \LaTeX\ Class Files,~Vol.~14, No.~8, August~2021}%
{Shell \MakeLowercase{\textit{et al.}}: A Sample Article Using IEEEtran.cls for IEEE Journals}


\maketitle

\begin{abstract}
Orographic soaring can significantly extend the endurance of micro aerial vehicles (MAVs), but circling behavior, arising from control conflicts between longitudinal and vertical axes, increases energy consumption and the risk of divergence. We propose a control switching method, named SAOS: Switched Control for Autonomous Orographic Soaring, which mitigates circling behavior by selectively controlling either the horizontal or vertical axis, effectively transforming the system from underactuated to fully actuated during soaring. Additionally, the angle of attack is incorporated into the INDI controller to improve force estimation. Simulations with randomized initial positions and wind tunnel experiments on two MAVs demonstrate that the SAOS improves position convergence, reduces throttle usage, and mitigates roll oscillations caused by pitch–roll coupling. These improvements enhance energy efficiency and flight stability in constrained soaring environments.
\end{abstract}


\section{Introduction}

The flight endurance of micro air vehicles (MAVs) significantly constrains operational capabilities, limiting the scope of missions they can perform \cite{karydis2017energetics,bronz2009towards}. This limitation is not solely due to inherently short flight durations, but also because take-off and landing procedures typically demand substantial time, energy, effort, and space. One potential solution to this problem lies in the advancement of battery technology, which could lead to improved efficiency. However, progress in this area has been relatively slow \cite{li2019practical, xu2023high}. Consequently, researchers have been exploring alternative solutions, such as using energy sources with higher energy densities or enabling mid-air refueling or recharging \cite{de2021nederdrone,mohsan2022comprehensive}. Nevertheless, these approaches require considerable investment in hardware and system infrastructure, and often necessitate larger, heavier platforms—undermining the fundamental advantage of MAVs being small.

An alternative approach is to exploit soaring, a flight technique widely employed by birds \cite{pennycuick1972soaring,bildstein1987hunting,strandberg2006wind} and human-piloted glider aircraft \cite{metzger1975optimal, akos2008comparing}. Soaring takes advantage of wind energy, specifically upward vertical winds, to gain altitude or remain airborne with minimal energy expenditure. A key strength of soaring is its compatibility with existing systems: it can be integrated into any fixed-wing aircraft without requiring hardware modifications, making it a valuable complement to other endurance-enhancing strategies. % By incorporating soaring techniques, overall energy consumption can be significantly reduced, enabling the potential for near-zero energy flight and, in some conditions, virtually indefinite endurance.

Various types of soaring techniques exist \cite{mohamed2022opportunistic}. Dynamic soaring is a technique that capitalizes on the wind's velocity gradient to gain energy, allowing migratory seabirds like albatrosses to traverse vast distances with minimal energy use \cite{weimerskirch2000fast,sachs2012flying}. This technique has also inspired research efforts to implement dynamic soaring in unmanned aerial vehicles (UAVs), aiming to extend their endurance by autonomously harvesting energy from wind gradients \cite{bonnin2015energy,mir2018review,park2025application}. Thermal soaring, exploiting updrafts generated by unevenly heated ground surfaces, is utilized by migratory birds and cross-country sailplanes \cite{harel2016adult,akos2010thermal,keskin2025adaptive}. Thermal soaring has been extensively studied in the context of control and guidance systems \cite{edwards2008implementation,cobano2013thermal,depenbusch2018autosoar}, as well as machine learning methods for autonomous MAVs \cite{reddy2018glider,flato2024revealing}.

\begin{figure}[tb]
\centering
\includegraphics[width=\linewidth]{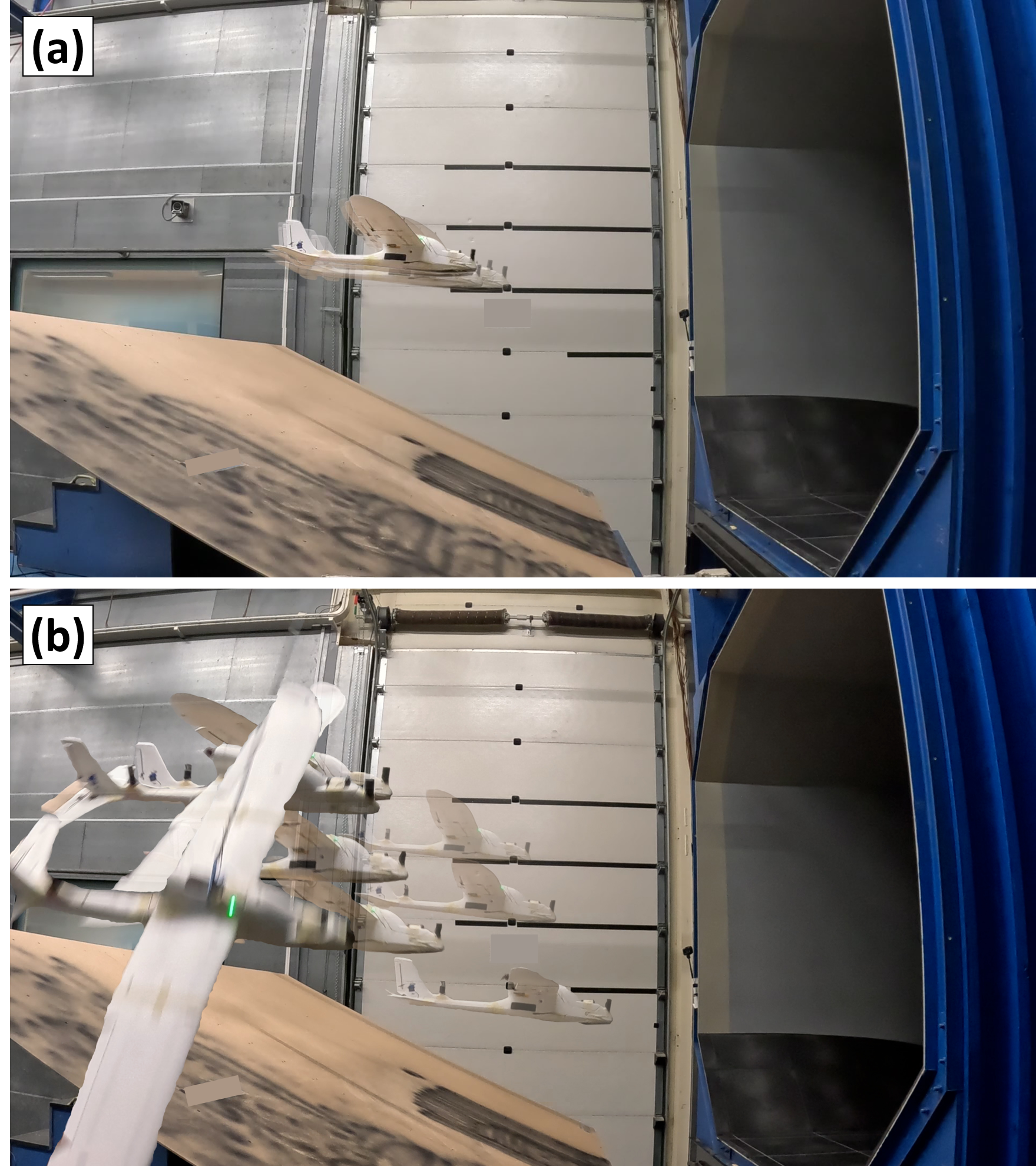}
\caption{
Wind tunnel test trajectories of the Seal plane over \textbf{10 seconds} using SAOS (proposed control switching method) and BASE (previous INDI-based controller
from AOSoar \cite{hwang2023aosoar} without any modifications) control methods.
(a) SAOS: minimal circling and stable soaring flight.
(b) BASE: pronounced circling behavior and trajectory divergence.
}
\label{f:main_pic_seal}
\end{figure}

Orographic soaring, another form of static soaring, utilizes updrafts created by natural or man-made obstacles such as dunes, mountains, or buildings \cite{bohrer2012estimating,sage2019orographic}. The Kestrel is a well-known example of a bird that employs this technique for hunting. Kestrels are often observed hovering in place over hills without flapping their wings as they search for prey—a behavior also referred to as wind-hovering \cite{videler1983intermittent,penn2022method}. 
%Although kestrels sometimes hover for more than a minute at a time (Brown and Amadon 1968), the mean duration of individual hovers is much less: 16-18 sec in American Kestrels (Bildstein 1978, Mills 1979) and 26 sec in Eurasian Kestrels (Shrubb 1982).

One of the key advantages of orographic soaring is its predictability. The obstacles that generate the updrafts are typically stationary, providing reliable and consistent lift. Orographic soaring also enables a fixed-wing MAV to hover in place. This wind-hovering not only helps extend flight distance and duration but also enhances surveillance capabilities. Thus, it offers potential for missions involving observation, surveillance, and even mid-flight recharging.

Orographic soaring has not been widely studied. Early research focused on analyzing wind fields, flight conditions, and the feasibility of soaring, including in urban environments using simulations and wind measurements \cite{white2012soaring, white2012feasibility, guerra2020unmanned, lu2025energy}. Real-world demonstrations with MAVs were performed, although they were limited by reliance on prior wind field knowledge and predefined trajectories \cite{fisher2015emulating}, or manual positioning before activating the soaring controller \cite{ de2021never}.
%Autonomous orographic soaring without wind field pre-knowledge has been shown, such as in front of a moving ship \cite{de2021never}, though manual identification of the soaring position was still required.
In wind tunnel environments, Suys et al. demonstrated autonomous orographic soaring with a glider MAV using a pitch controller \cite{suys2023autonomous}. While it required minimal wind-field knowledge, some manual setup—such as setting a target gradient line—was still needed.
Following that, AOSOAR achieved the first fully autonomous wind-hovering flight demonstration, requiring no human intervention, prior wind field knowledge, nor precise initial positioning, using a powered fixed-wing MAV \cite{hwang2023aosoar}. Utilizing an Incremental Nonlinear Dynamic Inversion (INDI) controller \cite{sieberling2010robust,smeur2020incremental} and an autonomous soaring position search method, it successfully performed wind-hovering flight with 0\% throttle for over 30 minutes continuously. The method demonstrated the ability to adapt to changing wind conditions without any prior wind-field data in the wind tunnel environment.

Although these studies have demonstrated autonomous soaring and wind-hovering, a specific flight pattern—circling—has been consistently observed. When the amplitude of the circling remains small, the MAV can still harvest a lot of energy from the updraft, though it incurs additional energy losses due to actuator activity and throttle adjustments. In more severe cases, the circling radius expands, causing the MAV to drift out of the updraft or into no-wind zones, which presents a significant challenge.
In wind tunnel environments, circling behavior could be reduced by controlling specific variables. Previous studies \cite{suys2023autonomous, hwang2023aosoar} showed that circling was reduced under certain combinations of wind speed and slope angle. However, despite these improvements, circling behavior often persisted. Moreover, in outdoor flights, manipulating the wind field is generally not feasible. Orographic soaring typically requires high wind speeds, which are often accompanied by strong gusts. These unpredictable conditions tend to worsen circling behavior and make stable soaring more difficult.

% our outdoor flight example
For example, during an outdoor flight over a sand dune, circling behavior was frequently observed and was more pronounced than in wind tunnel tests. Stalls occurred during circling, particularly when the MAV was flying backward. Although the MAV was sometimes able to recover its position and attitude, after several circling loops, it drifted out of the updraft and could no longer recover. Manual intervention by a pilot was required due to its proximity to the ground. This circling behavior was observed in both controlled wind tunnel environments and outdoor flights, indicating that it is a common phenomenon in orographic soaring.

% so we solve this problem???

\begin{figure}[b]
\centering
\includegraphics[width=\linewidth]{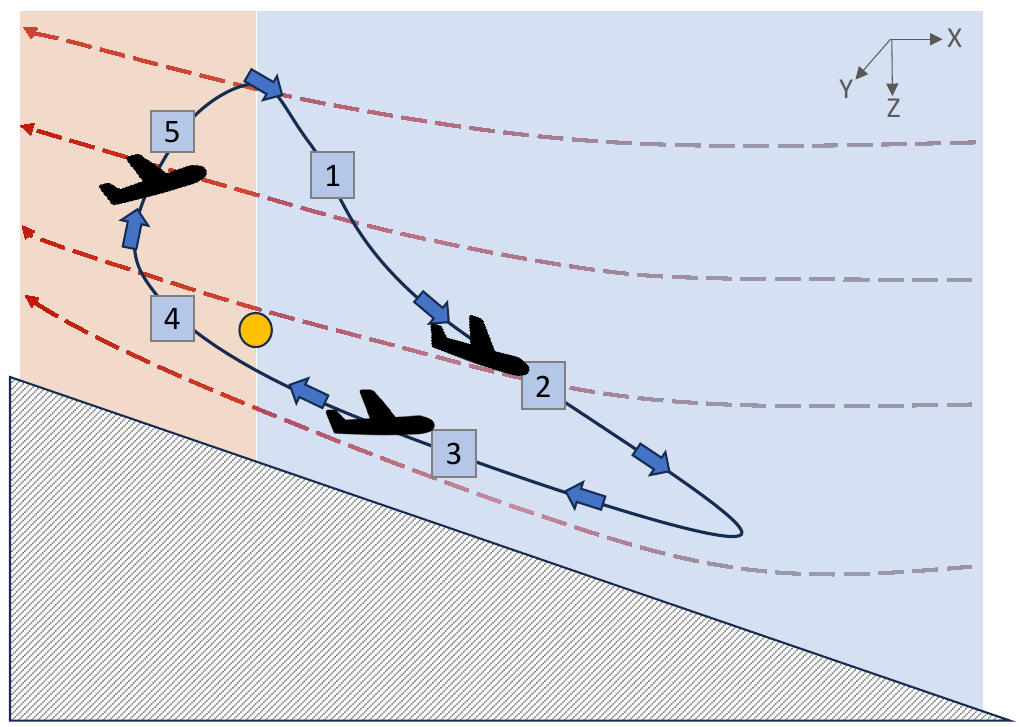}
\caption{
Illustrative example of MAV flight trajectory near a ramp during orographic soaring. Colored streamlines represent wind flow, with a gradient from gray to red indicating increasing wind speed. Airplane icons show the MAV’s position and pitch angle at different time steps along the path. The yellow dot marks the predefined reference position.
}
\label{f:circling_drawing}
\end{figure}

\begin{figure}[bht]
\centering
\includegraphics[width=\linewidth]{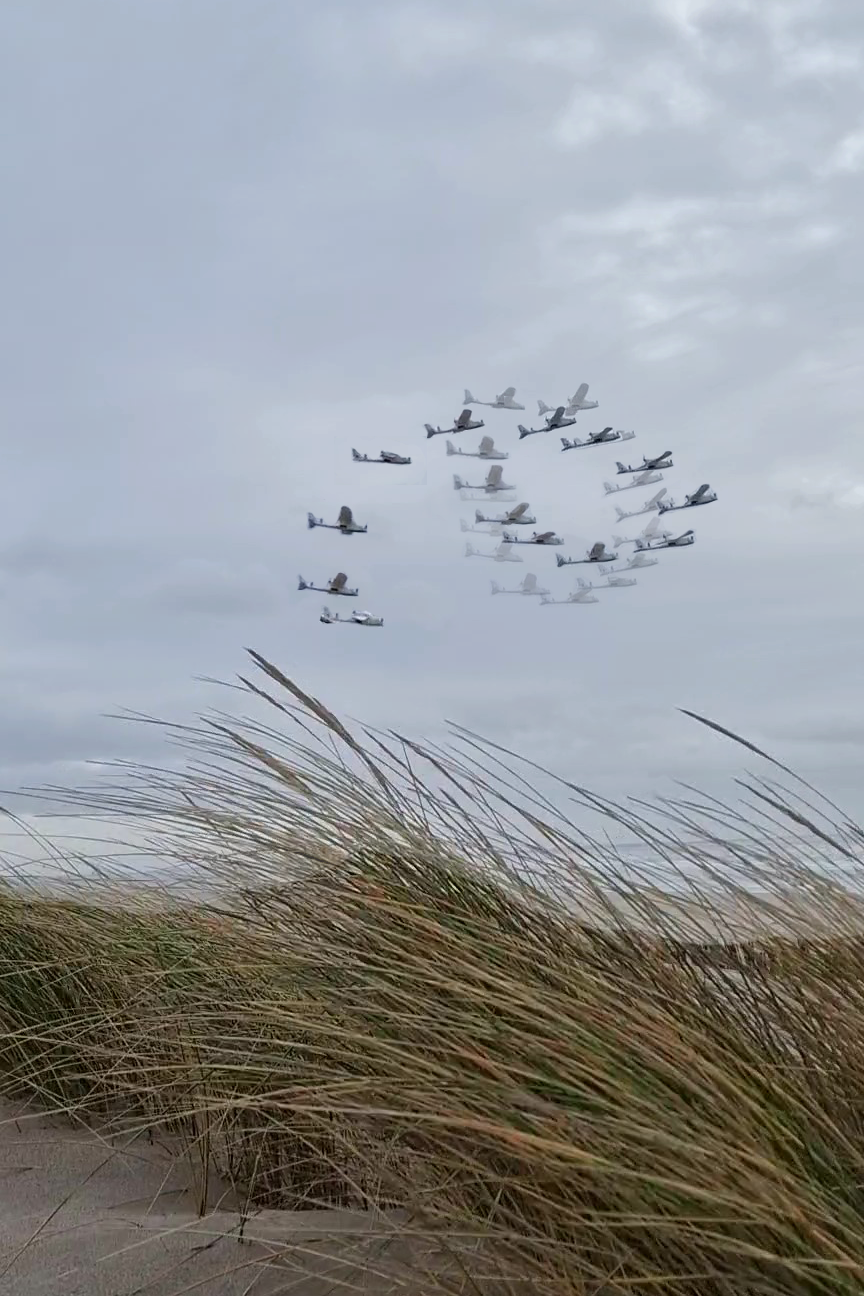}
\caption{Time-overlapped image extracted from outdoor flight footage of the Seal G-1500 MAV, demonstrating circling behavior over a sand dune.}
\label{f:circling_outdoor}
\end{figure}

% \begin{figure}[bth]
% \centering
% \includegraphics[width=\linewidth]{images/outdoor_stall_traj_placeholder_plots.JPG}
% \caption{Placeholder for the circling example}
% \end{figure}

% but there is no WHY here, or clear statement of the GOAL
% ------------------------------------------------------------
% contribution below

%achieved 'less' divergence, 'more' position accuracy or 'less' position error,
%'more' stable (*show statistical results)
%*so 'more' soaring overall using the proposed methods*

In this paper, we address a previously unexplored challenge in orographic soaring: \textit{\textbf{circling}}, a distinct flight behavior that prevents MAVs from maintaining stable wind-hovering. To mitigate this issue and enhance the robustness of soaring performance, we propose three key methods:

1. Angle of Attack (AoA) sensing: An AoA sensor is integrated to actively prevent stalls and regulate AoA through pitch control.

2. Enhanced aerodynamic modeling: An additional drag term is introduced, and AoA measurements are used to compute aerodynamic forces more accurately within the control model.

3. Control-mode switching: A control strategy is developed to transition the MAV from an underactuated to a fully actuated control mode during soaring, particularly when the throttle input is near zero.

In the remainder of the article, we will describe how these methods were implemented on MAV platforms and validated through successful wind-hovering demonstrations in a wind tunnel environment using two different MAV configurations.

%\textbf{TODO: refine/rewrite} \\
The remainder of this article is structured as follows:
Section 2 provides a detailed explanation of circling behavior in orographic soaring.
Section 3 describes the methods developed to mitigate this behavior.
Section 4 presents simulation results comparing different combinations of control strategies.
Section 5 outlines the experimental setup for the wind tunnel tests.
Section 6 presents the corresponding flight experiment results.
Section 7 offers a discussion of the findings and their implications.
Finally, Section 8 concludes the paper with a summary, final remarks, and suggestions for future work.

% Circling in depth
\section{Circling in Orographic Soaring}

Before delving into the proposed solutions to the circling behavior in orographic soaring, we first explain this problem in more detail.
Circling is a flight behavior observed in orographic soaring, where the MAV fails to converge to the intended soaring position but instead continuously overshoots it in the perpendicular axis, resulting in continuous circular or elliptical flight patterns around the target location. % This behavior typically arises when the MAV is near a feasible soaring region and consequently reduces its throttle to nearly zero.
The issue primarily arises from the inherent characteristics of the orographic wind field, where both vertical and horizontal wind speeds vary along the slope. A perturbation in one axis, therefore, also creates a perturbation in the other axis, which reduces the stability of soaring at a fixed position. While it is not specific to any particular controller, the controller can influence the behavior, either mitigating or exacerbating it.
To better understand this phenomenon, we present a comprehensive example that illustrates circling behavior in the context of orographic soaring.

\begin{figure}[t]
\centering
\includegraphics[width=\linewidth]{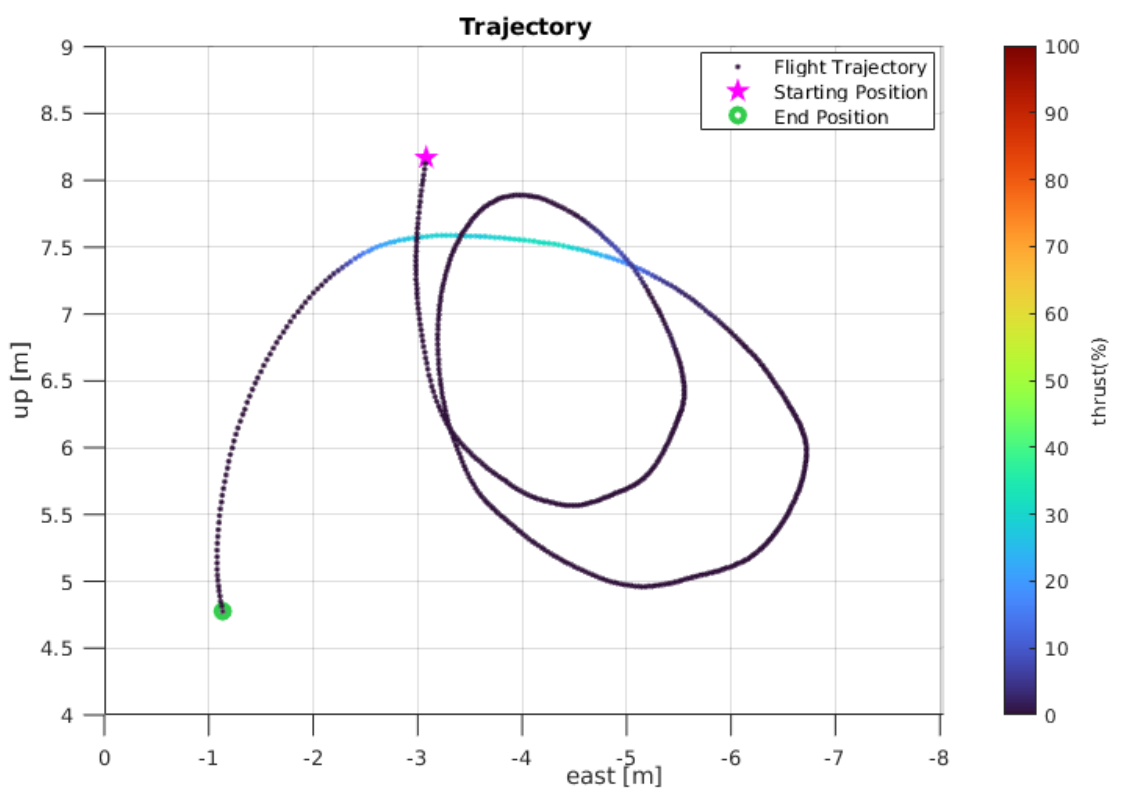}
\caption{Flight trajectory of the Seal G-1500 MAV exhibiting circling behavior during orographic soaring over a sand dune. The trajectory is reconstructed from position data recorded during an outdoor flight test.}
\label{f:circling_traj_plot}
\end{figure}

% difference between two circlings? --> controller setting/behaviour
% one is at high pitch (high altitude gain)

Circling behavior can occur in both indoor and outdoor environments.
We present two cases observed during an indoor test (figure \ref{f:main_pic_seal} and \ref{f:circling_drawing}) and an outdoor test (figure \ref{f:circling_outdoor} and \ref{f:circling_traj_plot}).
Depending on the controller settings, two distinct circling patterns emerged: horizontal and vertical. One was more prevalent indoors, while the other was more common outdoors, as parameter adjustments were made to accommodate the differences in available altitude and longitudinal space.

We first explain the circling behavior observed in the indoor wind tunnel experiments, illustrated in figure \ref{f:circling_drawing}, which arises when the longitudinal component dominates, typically due to high longitudinal position gains in the controller. In this scenario, the MAV exhibits more aggressive longitudinal movement relative to its altitude.

%This behavior is more common in indoor environments than outdoors, such as the TU Delft open-jet facility (OJF) wind tunnel, where the limited height of the wind section tightly constrains vertical motion. To ensure precise altitude control under these constraints and reduce the risk of overshooting, relatively low altitude gains were used. In contrast, the longitudinal space is relatively unrestricted and poses less risk when overshooting, allowing for greater exploratory motion along the longitudinal axis. Consequently, longitudinal control gains were set relatively higher.

%%
This behavior is more common in indoor environments than outdoors, such as the open-jet facility (OJF) wind tunnel, where the limited height of the wind section tightly constrains vertical motion. To avoid aggressive oscillations or overshooting under these tight constraints, relatively low altitude control gains were used, allowing smoother but slower vertical responses. In contrast, the longitudinal axis is relatively less constrained and more tolerant of overshooting, so higher longitudinal control gains were employed to better regulate forward and backward motion.
%%

% summary
The circling behavior occurs when the MAV attempts to approach a feasible soaring position, where the wind speed counteracts its airspeed and sink rate. However, due to its momentum, the MAV may slightly overshoot the reference position.
When, for instance, the MAV overshoots in the forward direction, it arrives in a section of the updraft with lower horizontal and vertical wind speeds, resulting in additional forward and downward acceleration. The forward overshoot then induces a small error in altitude. To counteract this, the controller commands a pitch-up maneuver, inducing both backward and upward acceleration. This response leads to a diagonal upward trajectory rather than a direct return to the reference point. It is worth noting that the longitudinal and vertical axes respond at different rates to pitch adjustments, making error correction challenging—especially in non-uniform wind fields such as updrafts.
When the MAV moves backward, it enters areas with stronger horizontal and vertical wind speeds, further accelerating it in the backward and upward directions. Upon reaching the reference position again, the MAV must cancel out the accumulated acceleration. However, with only the elevator available for control because the throttle is already zero while flying backward, achieving this correction instantaneously is difficult, possibly leading to another overshoot. 
As the MAV moves toward the rear side of the slope, where the updraft is stronger, it gains altitude. In this case, horizontal acceleration can also be corrected using throttle input. As a result, horizontal position errors are quickly mitigated, but altitude errors persist.
To reduce altitude error, the MAV pitches down; however, this induces forward acceleration, leading to yet another overshoot. Unlike in backward flight, where the throttle can be used to counteract motion, forward acceleration is more difficult to control. Consequently, the MAV overshoots the reference position again and the cycle repeats, resulting in a persistent circling pattern.

%% Original
The other circling case, shown in figure \ref{f:circling_outdoor}, occurs more frequently in outdoor environments. In such settings, controller configurations typically differ from those used indoors due to the availability of larger airspace and differing flight scenarios. 

%As a result, control gains are generally set higher for outdoor flights. For practical reasons, in outdoor operations, autonomous flights often begin from a higher altitude, and the vertical airspace tends to be less constrained than the horizontal space, which may be limited by obstacles or terrain features that generate updrafts. Accordingly, altitude gains are usually set relatively higher than longitudinal gains, which are set more conservatively.
%In this scenario, the controller responds more aggressively to altitude deviations, resulting in larger variations in pitch angle as it attempts to minimize vertical error. When the MAV overshoots the reference altitude and climbs too high, the controller commands a pitch-down maneuver to descend, causing the MAV to descend and move forward. As it descends, it overshoots the target altitude again, this time going too low. The controller then commands a pitch-up response to regain altitude, which drives the MAV upward and backward. This pattern repeats, leading to a persistent oscillatory behavior.
%Although the dynamics may look different from the indoor circling case, the underlying pattern remains the same: the MAV continuously overshoots the reference position due to coupling between pitch and longitudinal motion and gradients in the updraft field.
%% 
For practical reasons, autonomous flights in outdoor conditions often begin from higher altitudes, and the vertical airspace is typically less constrained than the horizontal space, which may be limited by terrain or obstacles generating updrafts. Accordingly, altitude control gains are often tuned more aggressively than longitudinal ones to maintain altitude in a wide vertical range.
However, this increased sensitivity can result in more reactive pitch behavior. When the MAV overshoots the reference altitude and climbs too high, the controller pitches down to descend, causing forward motion. After descending, it may overshoot again and go too low, prompting a pitch-up response that sends it backward and upward. This coupling between altitude and longitudinal motion leads to a repeating oscillatory pattern.
Although this behavior differs visually from the circling observed indoors, it stems from the same underlying issue: coupled dynamics between pitch and acceleration in a non-uniform wind field.
%%

%when the vertical component becomes dominant, for example, when the controller employs high altitude gains. In this scenario, the controller reacts more aggressively to altitude deviations, resulting in larger variations in pitch angle. This behavior is more common in outdoor environments, where ample airspace allows for greater altitude adjustments while longitudinal movement is more constrained by obstacles or terrain that limit access to updrafts. Additionally, outdoor operations typically require starting the soaring position search from a higher altitude due to practical takeoff limitations. As a result, the controller is configured to prioritize precise horizontal tracking while permitting faster altitude corrections.

It is important to note that both circling behaviors can occur under identical wind conditions and within the same environment using the same MAV platform, depending solely on the controller configuration.
In this paper, we mainly focus on the longitudinal position-dominant case. % * (OR are we going to explain both?)

While the MAV can technically continue to \textit{soar} during circling, this behavior introduces several critical issues:

1. Energy Inefficiency – Circling wastes energy, which contradicts the primary goal of soaring as an energy-efficient flight mode. Although the MAV can maintain flight with minimal power, some throttle input is typically required to counteract the backward acceleration that occurs during circling. Additionally, continuous engagement of control surfaces increases overall energy consumption.

2. Reduced Stability for Observation – Unlike wind-hovering, which enables the MAV to remain stationary for observation, circling prevents it from holding a fixed position. This limits its effectiveness for tasks such as surveillance, monitoring, or environmental data collection.

3. Increased Risk of Instability and Crashes – While the circling radius may remain stable, it can also expand unpredictably. Excessive divergence poses a risk of flyaway or collision, especially in orographic soaring scenarios where the MAV operates near terrain or obstacles to exploit updrafts. Drifting out of the updraft region may necessitate a recovery maneuver, which usually consumes a considerable amount of energy and effort, or, in the worst case, leads to a crash or fly-away.

To address these challenges, we propose methods to minimize circling and enable more stable orographic soaring. Reducing circling is essential to conserve energy, improve flight stability, and enhance overall safety.

\section{METHODOLOGY}

\begin{figure*}[bth]
\centering
\includegraphics[width=0.95\linewidth]{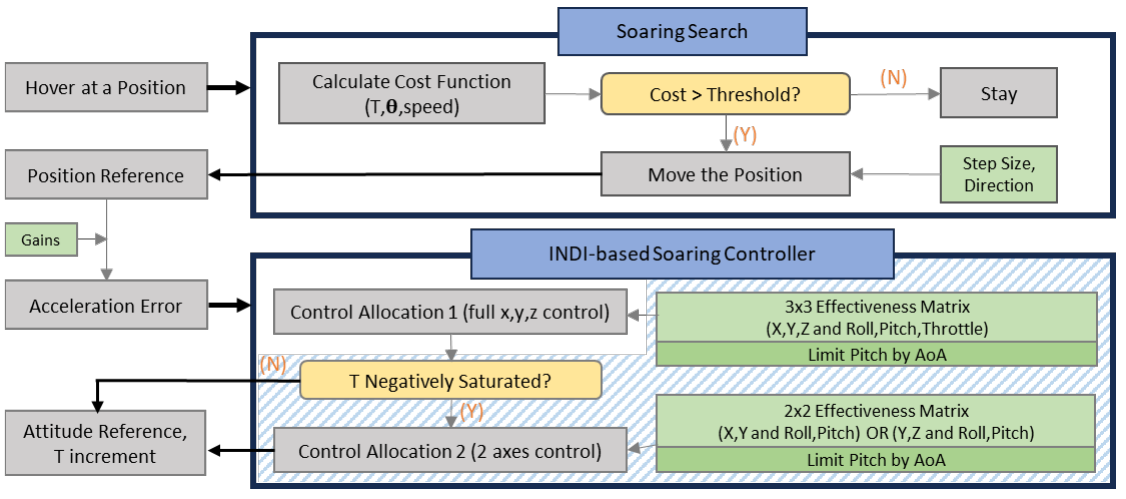}
\caption{The interaction between the soaring search algorithm and the INDI-based soaring controller. The diagonally striped area is newly introduced in this paper. In the controller, two control allocations are employed to handle cases where the thrust (T) is negatively saturated, switching to a reduced 2-axis control while limiting pitch by angle of attack (AoA).}
\label{f:ctrl_flow}
   % \vspace{-3mm}
\end{figure*}

% in terms of the controller, WHY those methods?
% drag, aoa, aoa effectiveness, switching 
% aoa limit -> prevent stall
% drag -> because of slow pitching up when flying backward (this goes into the aoa eff subsection)
% aoa effectiveness -> high error in the model (outer loop) due to pitch-aoa error
% switching -> underactuated system: hor,ver components conflict / handling model error

We introduce improvements to the soaring controller to prevent or reduce circling behavior, using the controller from AOSOAR\cite{hwang2023aosoar} as a baseline. 
AOSOAR has two main parts: the autonomous soaring search and the orographic soaring controller. We focused on improving the control part in this paper.
For the control part of AOSOAR, an INDI-based controller with a weighted mean square (WMS) control allocation was used. It enables the MAV to move forward or backward in the updraft, and also cut off the throttle whenever required without changing any settings in-flight. \\
Figure \ref{f:ctrl_flow} shows the flow chart of the soaring search and controller from AOSOAR, and the area with blue hatched lines indicates the newly introduced part in this paper.

The enhancements to improve the circling behavior address two key issues: model errors and control conflicts caused by the underactuated system. In this section, we present the methods used to tackle these challenges.

\subsection{Angle of Attack (AoA) sensor}
% Does this belong to the hardware??
We implemented an AoA sensor on our MAVs using a 3D-printed wind vane with a potentiometer to measure the angle. This was done by repurposing a servo actuator as a pure encoder by removing the electric motor, while the angle is still directly measured and transmitted to the autopilot system. Despite minimal friction, the weight of the wind vane and remaining friction imposed a minimum operational wind speed for the sensor, approximately 6.0$m/s$. Below this speed, measurements were inaccurate. However, since the minimum wind speed for soaring in both MAVs exceeds this threshold, the limitation had no significant impact.

\begin{figure}[bt]
\centering
\includegraphics[width=1.0\linewidth]{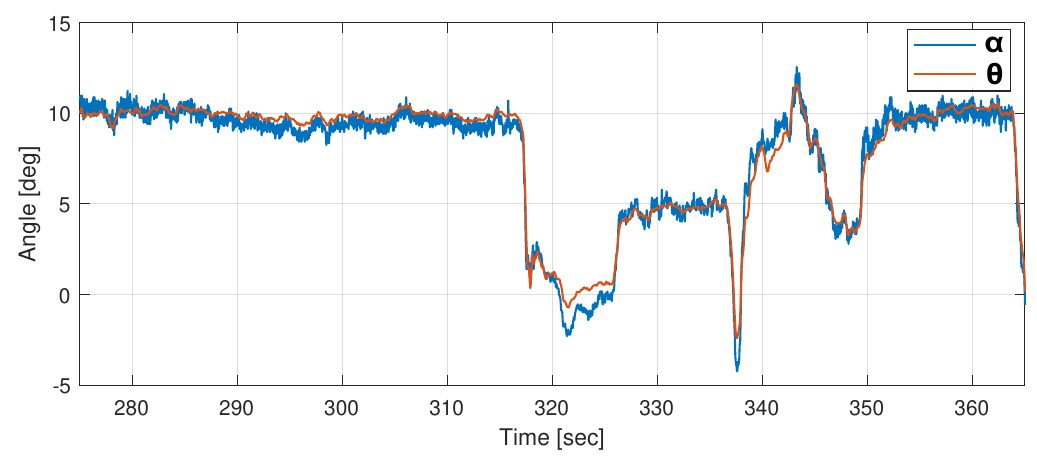}
\caption{Angle of attack ($\alpha$) measurement and pitch ($\theta$) angle recorded from flight log in a wind field without updraft, at a nominal wind speed of 10 m/s.}
\label{f:aoa}
   % \vspace{-3mm}
\end{figure}

The AoA sensor was manually calibrated in the wind tunnel. The testing result of the AoA sensor is shown in figure \ref{f:aoa}, yielding a root mean square error (RMSE) of 0.575 degrees and a maximum error of 2.118 degrees within the typical operational range for our soaring flight. %, which spans from -5 to 13 degrees.

%Due to space constraints, mounting the sensor near the wing caused disturbances in the wind measurement. As a result, the measured AoA did not perfectly align with the pitch angle due to aerodynamic effects. Compensation values were applied, resulting in a pitch error of X degrees within the operational range (-20° to 20°). An offset was introduced to ensure the AoA read 0 degrees at a 0-degree pitch.

\subsection{AoA regulation through pitch control}
With higher altitude or pitch gains, the MAV can stall due to a high angle of attack in the updraft. It can worsen the circling behavior, lead to divergence, or potentially cause a crash. Since the wind field is non-uniform and wind components change with position, preventing stall solely through pitch measurements is not feasible. To address this, we used the AoA measurement to regulate it by limiting the pitch angle. The maximum AoA was determined based on the stall angle estimated by experiments in the wind tunnel for each MAV platform, using:

\begin{equation}
% formula: max pitch increment = max aoa - current aoa
\theta_{max} = (\alpha_{max} - \alpha) + \theta
\end{equation}
where $\alpha$ is the angle of attack, and $\theta$ is the pitch angle.

\subsection{Outer loop control effectiveness with additional terms}

In AOSOAR\cite{hwang2023aosoar}, the drag term was omitted, and lift estimation was simplified, assuming zero flight path angle. However, updraft soaring introduces unique dynamics, such as backward flight and high angles of attack, requiring adjustments. To better model these conditions, we incorporated the drag term and utilized AoA for force calculations.

\begin{comment}
In the previous research, the effectiveness matrix was calculated based on the \cite{smeur2020incremental}.
Drag term was omitted in this study because it is relatively small in the normal flight conditions.
Lift derivatives were calculated based on a fitted function of pitch using flight log data. 
Lift estimation was also simplified, assuming small flight path angle.
However, soaring in the updraft has some difference from the normal level flight condition.
It can fly backward due to high horizontal wind, which is not very common for the normal fixed-wing flight.
It can also have high angle of attack because of the updraft, which is also not common for normal flight missions. Therefore we included the drag term in the effectiveness matrix, and used AoA to calculate forces.
\end{comment}

%% acc eq -> Taylor expansion -> partial derivatives -> (NED/body/wind rotations and assumptions: vector direction)
%% -> effectiveness matrix for L,D,T

The MAV's acceleration in the North-East-Down (NED) frame can be expressed as:
\begin{equation}
    \ddot \xi = g + \frac{1}{m}L_N(\eta,\alpha,V)+\frac{1}{m}D_N(\eta,\alpha,V)+\frac{1}{m}T_N(\eta,T)
    \label{eq:acc}
\end{equation}
Where $\ddot \xi$ is the second derivative of position, L is lift, D is drag, T is thrust, $\eta$ represents the attitude, $\alpha$ is the angle of attack, V is airspeed, g is the gravity vector, and m is the mass of the vehicle. 

The thrust vector in the NED frame is given by:

\begin{equation}
    T_N = M_{NB}T_B = 
    \begin{bmatrix}
        (c\theta c\psi - s\phi s\theta s\psi)T \\
        (c\theta s\psi + s\phi s\theta c\psi)T \\
        -c\phi s\theta T
    \end{bmatrix}
    \label{eq:Tn}
\end{equation}
Where c and s denote the cosine and sine functions, respectively.
For the lift and drag vectors, we assumed that the direction of the vectors does not rotate with the pitch or wind angles, in order to simplify the model. The amount of the lift and drag depends on the angle of attack, but it does not affect the direction of the vectors. The sideslip angle was assumed to be zero because, in the wind-hovering flight, the heading is aligned with the wind direction. Thus, the lift vector in the NED frame is given by:
\begin{equation}
\begin{split}
    L_N 
    % = M_{NB}M_{BW}L_W(\alpha,V) \\
    = M_{NB}^{\theta=0}M_{BW}^{\beta=0}L_W^{\alpha=0}(\alpha,V)
        = \begin{bmatrix}
        s\phi s\psi L(\alpha,V) \\
        -s\phi c\psi L(\alpha,V) \\
        c\phi L(\alpha,V)
    \end{bmatrix}
    % = \begin{bmatrix}
    %     (s\theta c\psi + s\phi c\theta s\psi)L(\alpha,V) \\
    %     (s\theta s\psi - s\phi c\theta c\psi)L(\alpha,V) \\
    %     c\phi c\theta L(\alpha,V)
    % \end{bmatrix}
\end{split}
\end{equation}

In the same manner, the drag vector in the NED frame is given by:
\begin{equation}
    D_N
    % = M_{NB}M_{BW}D_W(\alpha,V)
    = M_{NB}^{\theta=0}M_{BW}^{\beta=0}D_W^{\alpha=0}(\alpha,V)
        = \begin{bmatrix}
        c\psi D(\alpha,V) \\
        s\psi D(\alpha,V) \\
        0
    \end{bmatrix}
    % = \begin{bmatrix}
    %     (c\theta c\psi-s\phi s\theta s\psi)D(\alpha,V) \\
    %     (c\theta s\psi + s\phi s\theta c\psi)D(\alpha,V) \\
    %     -c\phi s\theta D(\alpha,V)
    % \end{bmatrix}
\end{equation}

Then, the control equation can be derived by taking partial derivatives by applying a Taylor expansion to the Equation \ref{eq:acc}:
% \begin{equation}
%     \ddot \xi = g + \frac{1}{m}L_N(\eta_0,V_0)+\frac{1}{m}D_N(\eta_0,V_0)+\frac{1}{m}T_N(\eta_0,T_0)
%     + \pdv{\phi}\frac{1}{m}L_N(\phi,\theta_0,\psi_0,V_0)|_{\phi=\phi_0}(\phi-\phi_0)
%     + ...
% \end{equation}
% It leads to the following equation:
\begin{equation}
    \ddot{\xi} \!=\! \ddot{\xi_0} \!+\! \frac{1}{m}\![G_T(\eta_0,T_0) \!+\! G_L(\eta_0,\alpha,V) \!+\! G_D(\eta_0,\alpha,V)]\!(u_c \!-\! u_{c0})
\end{equation}
where G is the control effectiveness matrix, and $u_c = \begin{bmatrix}    \phi & \theta & T \end{bmatrix}^T$.
The G matrices are given by:
\begin{equation}
\begin{split}
    G_T(\eta,T) = \\
    \begin{bmatrix}
        \!-\!c\phi s\theta s\psi T \!& \!(\!-\!s\theta c\psi\!-\!s\phi c\theta s\psi)T \!&\! c\theta c\psi\!-\! s\phi s\theta \!s\psi \\
        \!c\phi s\theta c\psi T \!& \!(\!-\!s\theta s\psi\!+\!s\phi c\theta c\psi)T \!&\! c\theta s\psi\!+\! s\phi s\theta c\psi \\
        s\phi s\theta T            \!&\!   \!-c\phi c\theta T                      \!& \!  -c\phi s\theta            
    \end{bmatrix}
    \end{split}
\end{equation}

%     $G_L(\eta,V) = \\$
% $    \begin{bmatrix}
%         c\phi c\theta s\psi L      &   (c\theta c\psi - s\phi s\theta s\psi)L + (s\theta c\psi + s\phi c\theta s\psi) L_d & 0\\
%         -c\phi c\theta c\psi L     &   (c\theta s\psi + s\phi s\theta c\psi)L + (s\theta s\psi - s\phi c\theta c\psi) L_d & 0\\
%         -s\phi c\theta L          &   -c\phi s\theta L + c\phi c\theta L_d    &   0
%     \end{bmatrix}$

\begin{equation}
        G_L(\eta,V) = 
    \begin{bmatrix}
        c\phi s\psi L(\alpha,V)      &   s\phi s\psi \pdv{\alpha}L(\alpha,V) & 0\\
        -c\phi c\psi L(\alpha,V)     &   -s\phi c\psi \pdv{\alpha}L(\alpha,V) & 0\\
        -s\phi L(\alpha,V)          &   c\phi \pdv{\alpha}L(\alpha,V)    &   0
    \end{bmatrix}
\end{equation}

%     $G_D(\eta,V) = \\$
% $    \begin{bmatrix}
%         -c\phi s\theta s\psi D      &  (-s\theta c\psi - s\phi c\theta s\psi)D + (c\theta c\psi - s\phi s\theta s\psi)D_d    & 0\\
%         c\phi s\theta c\psi D       &  (-s\theta s\psi + s\phi c\theta c\psi)D + (c\theta s\psi + s\phi s\theta c\psi)D_d  & 0\\
%         s\phi s\theta D          &    -c\phi c\theta D - c\phi s\theta D_d         &   0
%     \end{bmatrix}$

\begin{equation}
    G_D(\eta,V) = 
    \begin{bmatrix}
        0 & -c\psi \pdv{\alpha}D(\alpha,V)  & 0\\
        0 & s\psi \pdv{\alpha}D(\alpha,V)    & 0\\
        0 & 0 &  0
    \end{bmatrix}
\end{equation}

\begin{figure}[bth]
\centering
\includegraphics[width=1\linewidth]{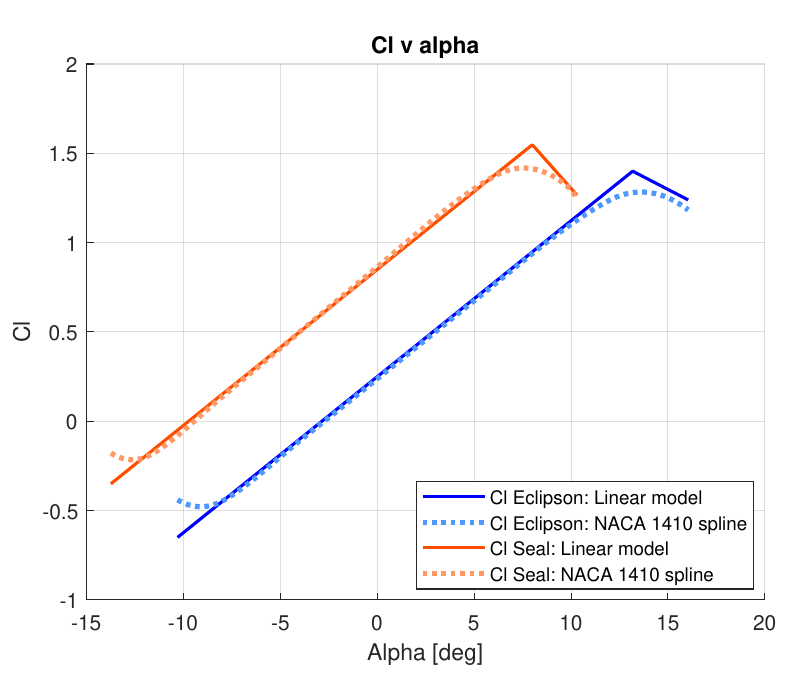}
\caption{
Lift coefficient (CL) versus angle of attack (Alpha) for the Eclipson C and Seal planes. The linear region was estimated from flight data, and the NACA 1410 airfoil model was converted to a spline to fit the estimated data, enabling stall-region modeling.
}
\label{f:cl_1410_plots}
   % \vspace{-3mm}
\end{figure}

Here, we assumed $\pdv{\theta}{\alpha}=1$. The wind flow is considered constant at a given position, and the wind direction is assumed to remain unchanged over time.
Lift and drag forces were estimated and modeled using flight-log data. However, accurately capturing the stall region was not feasible using flight data alone. Although a stall is ideally avoided during flight, incorporating this region into the model allows the controller to autonomously manage stall recovery. In practice, the aircraft may exceed the critical angle of attack due to turbulence, gusts, or variations in the wind field. Therefore, a realistic representation of stall behavior is essential, as opposed to abruptly dropping the lift force to zero.
To address this limitation, we estimated the stall region using a NACA 1410 airfoil model. 
The lift coefficient versus angle of attack is shown in figure \ref{f:cl_1410_plots}.
A base model was selected based on wing measurements and then adjusted through offsets and scaling to better fit the observed data. This approach enables the use of a standardized airfoil model with minimal parameter tuning, facilitating adaptation across different MAV platforms.

\subsection{Control switching method}

%1. three axes and two actuators (x,y,z and elev, ail) when thr == 0
%- x error and z error conflict with each other
%2. negatively saturated throttle gives further model error

%
Another challenge arises when the MAV must control its position along three axes (longitudinal, lateral, and vertical) but has only two available actuators, an elevator for pitch and ailerons for roll, particularly in the absence of throttle input.
This situation typically occurs when the MAV has too much total energy near the feasible soaring position.
The feasible soaring position refers to a location where the MAV can, in theory, perform wind-hovering, where the wind is sufficient to keep the MAV aloft without requiring thrust. \\As the MAV approaches such positions, it gradually reduces throttle input, as additional thrust would push it away from the desired location, and tries to soar.
When the throttle becomes 0\%, the MAV adjusts pitch to minimize errors in both longitudinal and vertical axes. However, pitch adjustments affect both axes simultaneously, creating conflicts. Pitching down increases forward acceleration but also induces downward motion, while pitching up increases backward acceleration but also induces upward motion. Since the two axes respond at different rates, these coupled effects make error corrections in both axes conflict with each other.

To address this issue, we implemented a control-switching strategy. Instead of attempting to control all three axes simultaneously, the controller selectively manages either longitudinal or vertical position in conjunction with lateral position, ignoring one axis based on predefined settings. %This approach enables more effective use of the two available actuators: the elevator and ailerons. 
This approach solves the underactuation problem between the longitudinal and vertical axes when the throttle is already at 0\% and pitch is the only available input for controlling both directions.

\begin{figure}[tb]
\centering
\includegraphics[width=1\linewidth]{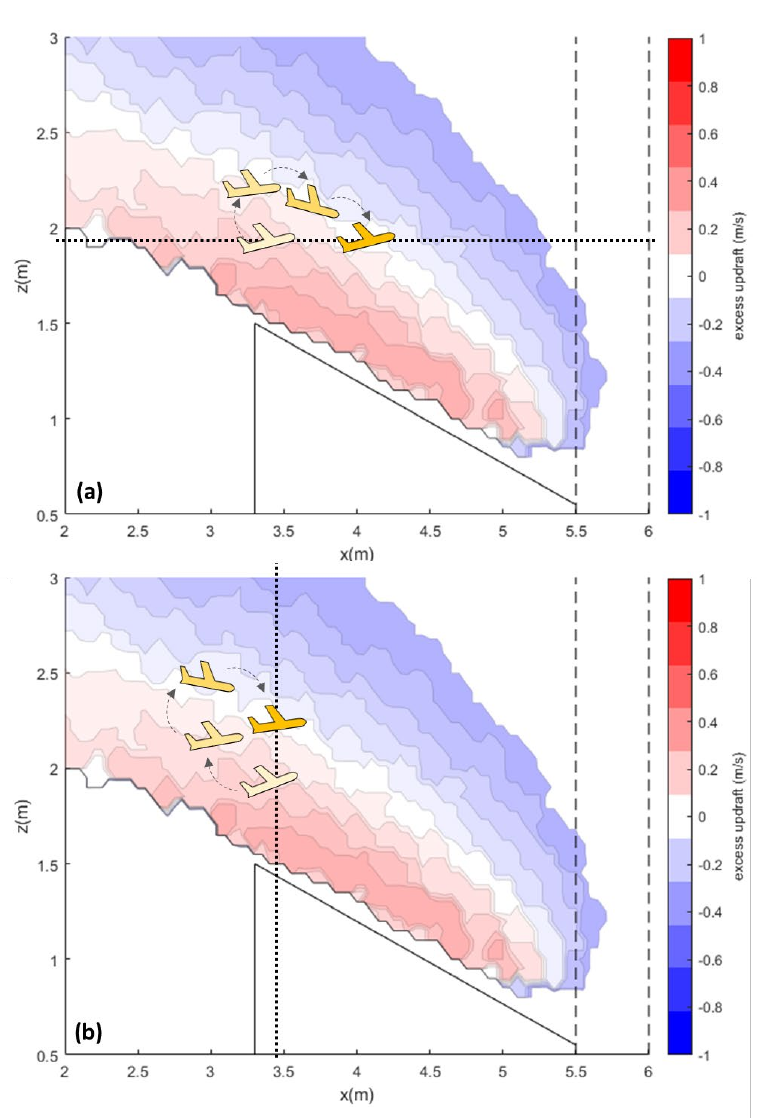}
\caption{
Examples of the OJF environment with excess updraft contours. The white region indicates the feasible soaring area, where the MAV’s sink rate is balanced by the updraft. Red areas have excessive updraft, while blue areas lack sufficient updraft to soar.
(a) Longitudinal axis controlled; vertical axis left uncontrolled.
(b) Vertical axis controlled; longitudinal axis left uncontrolled.
}
\label{f:drawing_sw_hv}
   % \vspace{-3mm}
\end{figure}

By relinquishing control of one axis—either longitudinal or vertical—the MAV can passively converge to the feasible soaring area, where its sink rate is balanced by the updraft.
Figure \ref{f:drawing_sw_hv} illustrates examples of this behavior, two cases are shown: (a) where only the vertical axis is controlled, and (b) where only the longitudinal axis is controlled.

When only altitude is controlled and the longitudinal position is left uncontrolled, the MAV responds to vertical disturbances autonomously. If the updraft is too strong, the MAV is naturally pushed upward. To correct this and maintain its target altitude, the MAV pitches down and begins to move forward to a region with weaker updrafts. If too low, the resulting pitch correction will make the MAV fly slower, moving backwards towards regions with stronger updrafts. Over time, this cycle continues until the MAV stabilizes in a region where the upward wind matches its natural sink rate—achieving an equilibrium that allows it to maintain position without active control in the longitudinal axis.
A similar principle applies when the longitudinal position is controlled while the altitude is left uncontrolled. If the updraft is too strong, the MAV is naturally pushed upward; if it is too weak, the MAV sinks. By actively maintaining its longitudinal position through control inputs, the MAV can passively adjust its altitude according to the available updraft. This is made possible by the natural tendency of updrafts to decrease with altitude. Over time, this allows the vehicle to settle at an equilibrium altitude—where the upward wind force balances its sink rate—without direct altitude control.

On the other hand, when the MAV is outside the soaring region or navigating toward a feasible soaring position, full three-axis control is required.
Therefore, we applied a cascaded control allocation method that dynamically adjusts the control model in response to throttle saturation. The controller initially computes a standard three-axis control allocation. If the throttle becomes zero or is negatively saturated, a secondary control allocation is triggered using a reduced model that limits control to two degrees of freedom. Crucially, this approach does not permanently disable the throttle term in the controller; instead, it temporarily switches the outer-loop control logic as needed. The MAV retains the ability to re-engage throttle input whenever it is necessary.
This strategy helps prevent the MAV from drifting too far backward beyond the reference position—a potentially unsafe condition—while preserving adaptability to wind gusts and environmental disturbances.

When the switching method is on, we use only two-axis components from the control effectiveness matrix. Depending on what axis to control, we can use two different settings: $G_{reduced_{xy}}$ is used to control longitudinal position instead of altitude, and $G_{reduced_{yz}}$ is used to control altitude instead of longitudinal position. In both settings, the lateral axis remains controlled.

\begin{equation}
    G_{reduced_{xy}} \!=\! 
    \begin{bmatrix}
        \!-\! c\phi s\theta s\psi T & (\!-s\theta c\psi \!-\! s\phi c\theta s\psi)T\\
        \!+\! c\phi s\psi L(\alpha,V) &  \!+\! s\phi s\psi \pdv{\alpha}L(\alpha,V) \\
        & \!-\!c\psi \pdv{\alpha}D(\alpha,V) \\\\
        
        c\phi s\theta c\psi T & (\!-\! s\theta s\psi \!+\! s\phi c\theta c\psi)T\\
        \!-\! c\phi c\psi L(\alpha,V) &  \!-\! s\phi c\psi \pdv{\alpha}L(\alpha,V) \\
        & \!+\! s\psi \pdv{\alpha}D(\alpha,V)\\
    \end{bmatrix}
    \label{eq:Gxy}
\end{equation}

% \begin{equation}
%     G_{reduced_{xy}} = 
%     \begin{bmatrix}
%         \!-\!c\phi s\theta s\psi T + c\phi s\psi L(\alpha,V) \!& \!(\!-\!s\theta c\psi\!-\!s\phi c\theta s\psi)T + s\phi s\psi \pdv{\alpha}L(\alpha,V) -c\psi \pdv{\alpha}D(\alpha,V) \\
%         \!c\phi s\theta c\psi T -c\phi c\psi L(\alpha,V) \!& \!(\!-\!s\theta s\psi\!+\!s\phi c\theta c\psi)T + -s\phi c\psi \pdv{\alpha}L(\alpha,V) + s\psi \pdv{\alpha}D(\alpha,V)\\
%     \end{bmatrix}
% \end{equation}

\begin{equation}
    G_{reduced_{yz}} = 
    \begin{bmatrix}
        \!(\!-\!s\theta s\psi\!+\!s\phi c\theta c\psi)T & \\
        \!-\!s\phi c\psi \pdv{\alpha}L(\alpha,V) & c\theta s\psi\!+\! s\phi s\theta c\psi \\
        \!+\! s\psi \pdv{\alpha}D(\alpha,V) \!&\!  \\\\
        \!-c\phi c\theta T & \\ 
        \!+\! c\phi \pdv{\alpha}L(\alpha,V) & -c\phi s\theta\\
        \!+\! s\psi \pdv{\alpha}D(\alpha,V)  \!& \!             
    \end{bmatrix}
    \label{eq:Gyz}
\end{equation}\\
% *
% $\theta = 0$ assumption (aerodynamic angle does not change)\\
% Lift and drag are functions of $\eta$ and $\alpha$\\
% Partial derivatives -> $\pdv{\alpha}{\theta} = 1$ (wind flow is constant at the position)
% - direction does not change, but magnitudes change depending on the AoA
% - Final G matrix is Gl+Gd+Gt
% *

%By integrating this adaptive control strategy, the MAV achieves greater stability and responsiveness, ensuring more efficient and reliable orographic soaring.

% add
%In this way, the MAV can automatically converges into a new equilibrium position. (*fig wind field example) OR keep the altitude more efficiently? 
%automatically .. because there is less updraft forward and too much updraft backward and by keeping the altitude .. automatically fly backward.. because at the front there is less updraft so more pitch up, going backward then gains more updraft so less pitch and slow down the horizontal acceleration.
%Note that it will not work if there is not enough wind but the soaring concept does not work in that case anyway..

% effectiveness mat (reduced)

\section{SIMULATION}

We implemented the new control effectiveness model and switching method in our MATLAB simulation environment. The environment features a wind field that was precomputed using ANSYS Fluent \cite{ansysfluent}, based on the geometry of the Open Jet Facility (OJF). %TU Delft Open Jet Facility (OJF).

To evaluate the effectiveness of the control switching method, we tested six different configurations, gradually adding improvements to the base controller from AOSoar. These cases were designed to assess the impact of various modifications, including angle of attack limitations, inclusion of the drag term, and the control switching method. The tested cases were as follows:
\begin{enumerate}
\item \textbf{\textit{BASE}} Controller: The original INDI-based controller from AOSoar \cite{hwang2023aosoar} without any modifications. %(\textbf{\textit{AOS}})
\item \textbf{\textit{AOS-A}}: BASE + AoA Limitation. Added a maximum AoA constraint and regulated AoA limiting pitch angle to prevent stall conditions.
\item \textbf{\textit{AOS-D}}: BASE + Drag Term Inclusion. Integrated the drag term into the control effectiveness matrix of the outer-loop controller. In this case, the pitch angle was used to estimate the forces. % Initially, the drag term was omitted since it is negligible in standard flight conditions. However, in soaring flight—where backward movement occurs—accounting for drag ensures the controller accurately models aerodynamic forces. \\
\item \textbf{\textit{AOS-E}}: BASE + Effectiveness Matrix Using AoA. Recalculated the outer-loop effectiveness matrix using AoA instead of pitch angle to provide a more accurate force estimation. %The original approach assumed a small difference between pitch and AoA, but soaring flights often exhibit significant variations.\\
\item \textbf{\textit{AOS-SW}}: BASE + Control Switching method. Applied the control switching method to the BASE controller while still using pitch angle for force calculations, without the drag term or AoA limitation. 
\item \textbf{\textit{AOS-E-SW}}: AOS-E + Control Switching method. Implemented the control switching method alongside the enhanced effectiveness model, which includes AoA constraint, the drag term, and force calculations based on AoA in the effectiveness matrix. This configuration integrates all the improvements that were previously discussed. 
\end{enumerate}

\begin{figure}[tb]
% \centering
\includegraphics[width=0.95\linewidth]{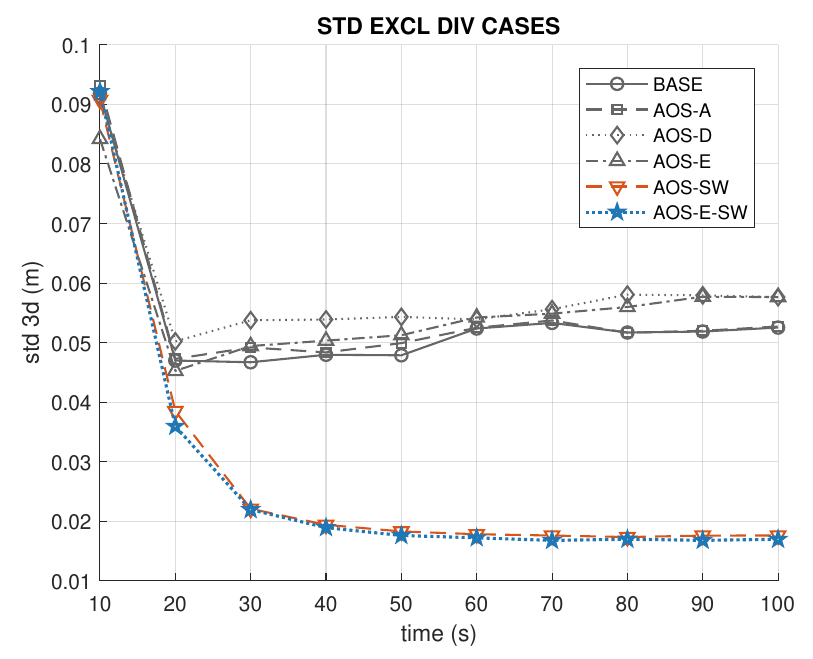}
\caption{
Position standard deviation (STD) over time for each control configuration. The plot shows the evolution of 3D position variability across non-diverged runs from 300 randomized reference positions. Lower STD values indicate greater soaring stability. Diverged cases were excluded to ensure a fair comparison; the convergence and divergence rates for each configuration are provided in Table \ref{tab:sim_rates}.
}
\label{f:sim_result_plot_pos}
   % \vspace{-3mm}
\end{figure}

\begin{figure}[tb]
% \centering
\includegraphics[width=0.95\linewidth]{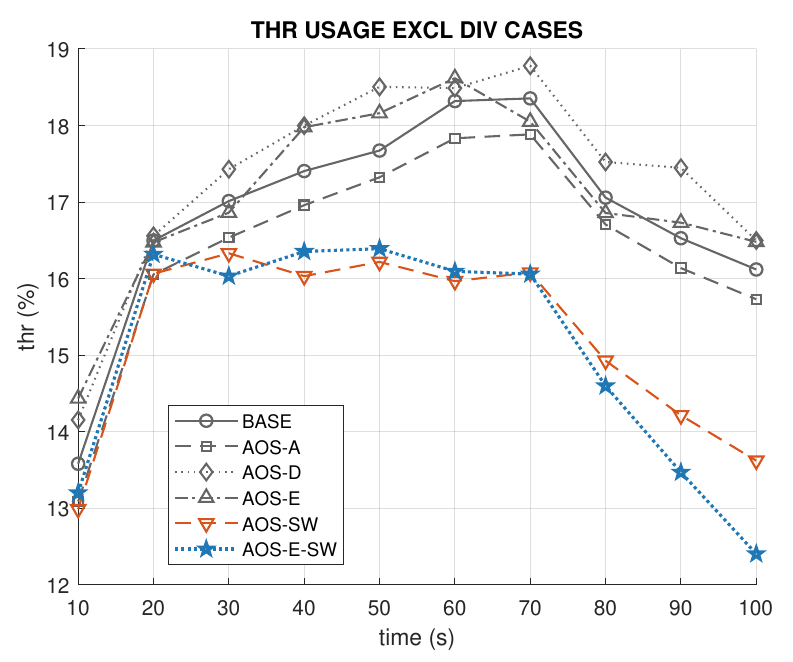}
\caption{
Mean throttle usage over time for each control configuration. The plot shows the average throttle input across non-diverged simulations from 300 randomized reference positions. Diverged cases were excluded from the analysis to ensure a fair comparison.
}
\label{f:sim_result_plot_thr}
   % \vspace{-3mm}
\end{figure}

% trajectory comparison #1 BASE and #6 ALL
\begin{figure}[bth]
% \centering
\includegraphics[width=0.95\linewidth]{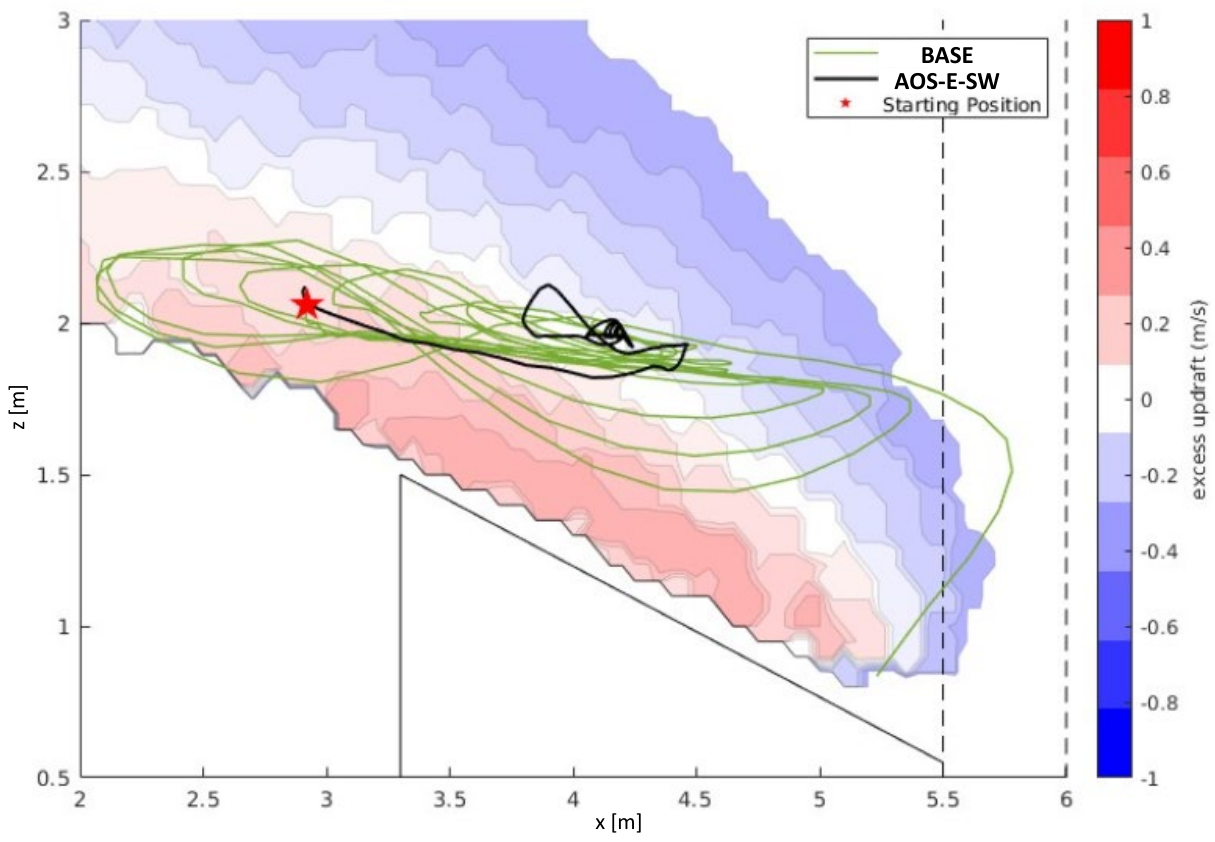}
\caption{
Trajectory comparison between the BASE and AOS-E-SW control configurations. The background contour indicates excess updraft: red regions represent overly strong updrafts, blue regions insufficient lift, and white regions indicate feasible soaring zones. The MAV using the BASE controller exhibited circling behavior and eventually diverged, while the AOS-E-SW configuration successfully converged to a feasible soaring position and maintained wind-hovering.
}
\label{f:sim_result_plot_traj}
   % \vspace{-3mm}
\end{figure}

Since we propose multiple modifications, we performed experiments with various configurations of the algorithm. These simulation experiments allow us to piece apart the specific contributions of the individual modifications.
Each control configuration was tested using the same set of 300 pre-generated reference positions. The MAV was commanded to hover at each reference position and was allowed to use the throttle as necessary. For instance, if the updraft was insufficient and the horizontal wind was too strong to hover at the position, the MAV could maintain its position by using the throttle. However, if the MAV stalled, the trial was marked as a divergence.
%However, if the wind conditions were too weak for the MAV to stay afloat and therefore stalled, the trial was marked as a divergence. 
Similarly, if the MAV began circling and exited the wind field due to position oscillation, the case was also classified as a divergence.

These simulations enabled a systematic evaluation of each control modification and validated the effectiveness of the control-switching method in improving soaring performance. The simulation was conducted at 300 randomly generated reference positions under a wind field calculated using ANSYS Fluent with a nominal wind speed of 7 $m/s$, with each run lasting 100 seconds. To eliminate transient effects, the first 10 seconds of data were excluded from the analysis. The remaining 90 seconds were divided into 10-second time windows, and within each window, the standard deviation (STD) of the MAV’s 3D position and the mean throttle usage were computed.

Figures \ref{f:sim_result_plot_pos} and \ref{f:sim_result_plot_thr} present the results for each control configuration. A case was classified as a divergence if the position STD exceeded 0.5 $m$, which reflects loss of position control beyond the typical bounds of the wind field, and as a convergence if the position STD was below 0.04 $m$ and the throttle usage remained under 3\%, corresponding to stable hovering within a narrow margin. Diverged cases were excluded from the statistical analysis to ensure a fair comparison, as their large values could skew the overall results.

The findings indicate that the control switching method was the most effective improvement. While other methods have only a little difference from the BASE method, the addition of the control switching method led to a significant improvement in position deviation. It played a crucial role in helping the aircraft converge to the soaring position and preventing divergence. The results demonstrate a higher convergence rate and reduced position deviation, leading to more successful soaring flights. %The success rate was notably higher with the control switching method. 

%AOS-SW and AOS-E-SW showed minimal differences, though AOS-E-SW performed slightly better statistically, showing an average position STD of 0.0176 and mean throttle of 13.6\% for AOS-SW and an average position STD of 0.0170 and mean throttle of 12.4\% for AOS-E-SW.

Among the tested configurations, AOS-SW and AOS-E-SW performed similarly, with AOS-E-SW achieving slightly better results. Specifically, AOS-SW achieved an average position standard deviation of 0.0176 and a mean throttle usage of 13.6\%, while AOS-E-SW further reduced the average standard deviation to 0.0170 and throttle usage to 12.4\%.

Based on these results, we selected AOS-E-SW as the final configuration for experimental testing and named the method \textbf{\textit{SAOS}}: Switched Control for Autonomous Orographic Soaring.

Figure \ref{f:sim_result_plot_traj} shows the trajectories for the BASE controller (Case 1) and SAOS (AOS-E-SW, Case 6). Under the BASE controller, the aircraft exhibited oscillatory behavior along the x-axis and eventually diverged. In contrast, SAOS enabled the aircraft to converge to and maintain the soaring position.

\begin{table}
    \centering
    \begin{tabular}{ccccccc}
         &  1&  2& 3 & 4 &  5&6 \\
         Conv. rate &  13.7\% & 13.3\% & 12.0\% & 14.7\% & 33.3\% & 33.7\% \\
         Div. rate & 41.3\% & 42.0\% & 43.0\% & 38.0\% & 33.7 & 35.7\% \\
    \end{tabular}
    \caption{Rates of convergence and divergence in soaring from 300 randomly selected starting points}
    \label{tab:sim_rates}
\end{table}

\section{EXPERIMENTAL SETUP}

To verify the simulation results, we conducted real-world experiments using two different MAVs in a wind tunnel environment. Below, we describe the environment setup, the MAV hardware used, and the experimental procedure.

\subsection{Wind tunnel setup}

The test environment was set up in the Open Jet Facility (OJF), which features a wind section measuring $2.85m \times 2.85m$. To generate an updraft, a ramp was installed in front of the wind tunnel outlet. The ramp measured $3.6m \times 2.4 m$, with a width greater than that of the wind section to help reduce downdrafts along the sides, and was inclined at an angle of 32 degrees. An Opti-Track system was installed and calibrated prior to testing to track the MAV’s position accurately. For safety, a ceiling-mounted rope system was used to prevent flyaways or crashes.

\subsection{MAV Hardware}
\begin{figure}[tb]
    \centering
    \includegraphics[width=1.0\linewidth]{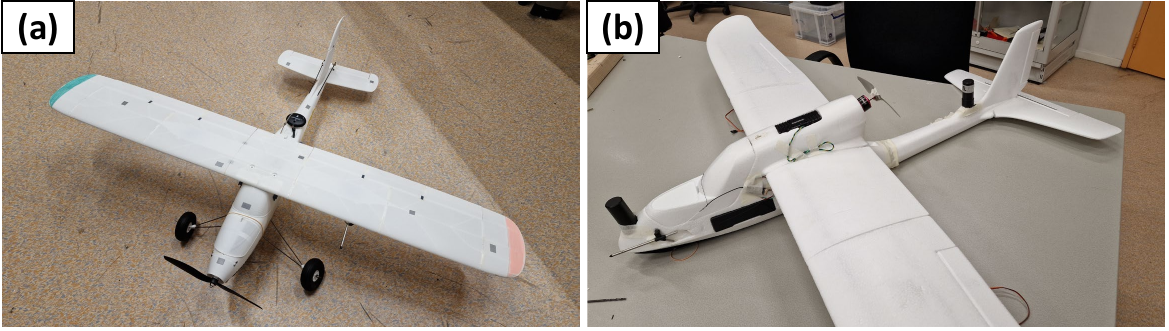}
    \caption{Two MAV platforms used for the wind tunnel flight tests: (a). Eclipson model C and (b). Seal G1500.
    }
    \label{f:mavs}
       % \vspace{-10mm}
\end{figure}

Two platforms of different sizes and configurations were used for testing to evaluate the generality of the proposed control method. The first was the Eclipson Model C \cite{eclipson}, shown in figure \ref{f:mavs} (a)., %the same model used in our previous research \cite{hwang2023aosoar}, 
featuring a wingspan of $1100mm$, a wing surface area of $18 dm^2$, and a total weight of $716g$. The second platform was the Eachine \& ATOMRC Seal G1500, shown in figure \ref{f:mavs} (b)., with a wingspan of $1500mm$, a wing area of $25.7 dm^2$, and a total weight of $1210g$. Both MAVs were equipped with an elevator, rudder, and ailerons for control, along with a single electric motor for throttle.

The electronics setup was similar across both platforms. We used the Paparazzi autopilot \cite{hattenberger2014using} running on a Pixhawk 4. Airspeed and angle-of-attack sensors were installed in different locations on each MAV to accommodate differences in available space and airframe design. Additionally, Opti-Track markers were attached to both MAVs for accurate indoor position tracking.

\begin{comment}
Merk: Eachine & ATOMRC
Model: Seal Wing G1500
Spanwijdte: 1500mm/1100mm
Lengte: 950mm
Materiaal: EPO
WingOppervlakte: 25.7d㎡/17.8dm²
Testparameters (De specificaties worden getest met PNP-versie: 2212/1400KV, 8040 prop, 21700/4s1p 4500mAh):
Aanbevolen startgewicht: 1100g (1500mm)
Maximaal startgewicht: 1400g (1500mm)
\end{comment}

\begin{figure}[tb]
    \centering
    \includegraphics[width=1.0\linewidth]{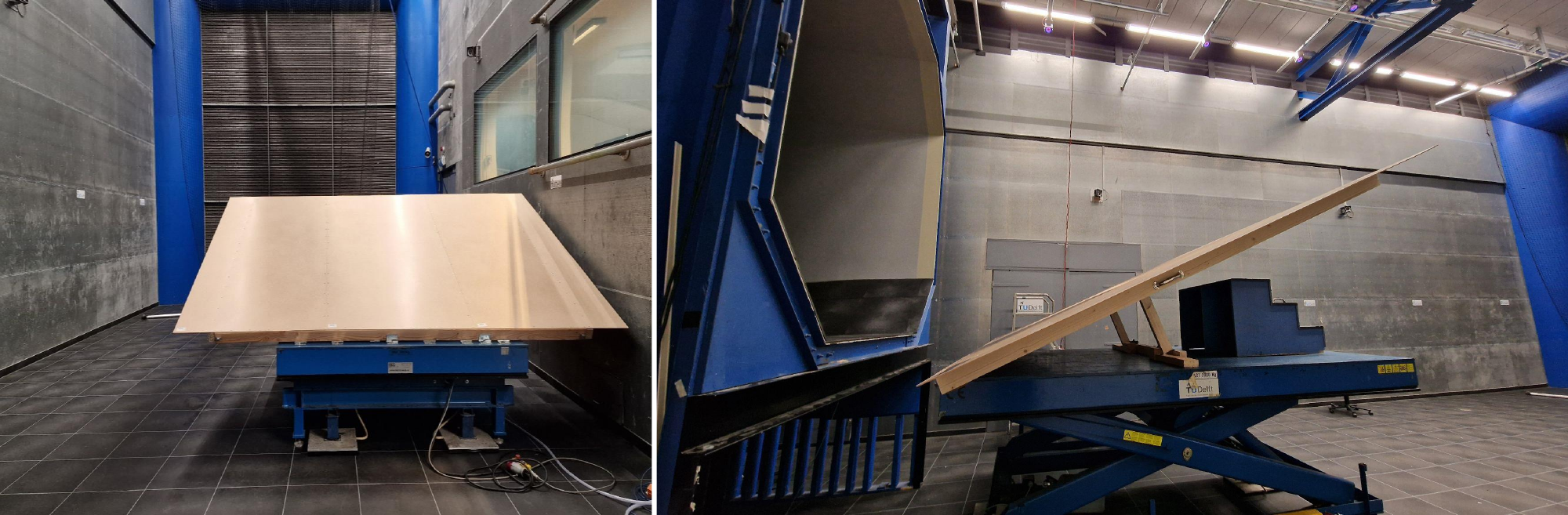}
    \caption{Ramp used to generate updraft (left) and overall test setup, including the wind section, at the Open Jet Facility (right).% TU Delft Open Jet Facility (right).
    }
    \label{f:ojf_setup}
       % \vspace{-10mm}
\end{figure}

\subsection{Test Procedure}

With the ramp installed, each flight began with manual launch and control of the MAV until the wind speed reached the target level, ensuring safe conditions for transition. Once stabilized, the MAV was switched to autonomous flight mode. To induce circling behavior, the reference positions were preset to place the MAV near a feasible soaring region.
After the MAV reached the target position, we alternated between the BASE controller and the controller switching method with additional components (SAOS) at the same position to observe and compare the circling behavior associated with each controller. Throughout the experiment, all parameters—including position gains and actuator effectiveness—were kept constant to ensure fair comparison. Flights were terminated if the MAV drifted out of the wind region and was unable to recover its position.
This experimental procedure was applied to both aircraft. Although the ramp inclination remained fixed, the nominal wind speed was adjusted for each platform to meet its specific wind and updraft requirements for soaring.
% 9m/s and 11m/s

% \begin{figure*}[htb]
% \centering
% \includegraphics[width=0.95\linewidth]{images/24_06_08__00_10_02_SD.jpg}
% \caption{Placeholder for the test results}
%    % \vspace{-3mm}
% \end{figure*}

\section{RESULTS}
%We first flew the Eclipson C model plane because we had previously achieved successful soaring flights with the same aircraft. While we used similar settings, note that there were some changes in the hardware and software configurations, so it does not have the same center of gravity, weight distribution, and parameters as the previous research, which impacts the soaring performance. Using the same test procedure, we repeated the same flight tests with another MAV.

In this section, we present the experimental results from the wind tunnel test.
Flight results from both platforms showed that the SAOS reduced position oscillations and throttle usage, allowing the MAV to maintain a steadier soaring flight. Switching back to the BASE controller from the SAOS immediately increased oscillations and position errors, indicating its difficulty in handling control conflicts. Overall, the results confirm that the SAOS effectively combats the circling behavior in soaring flight. 

\subsection{ECLIPSON C}
\begin{figure*}[bth]
\centering
\includegraphics[width=1.0\linewidth]{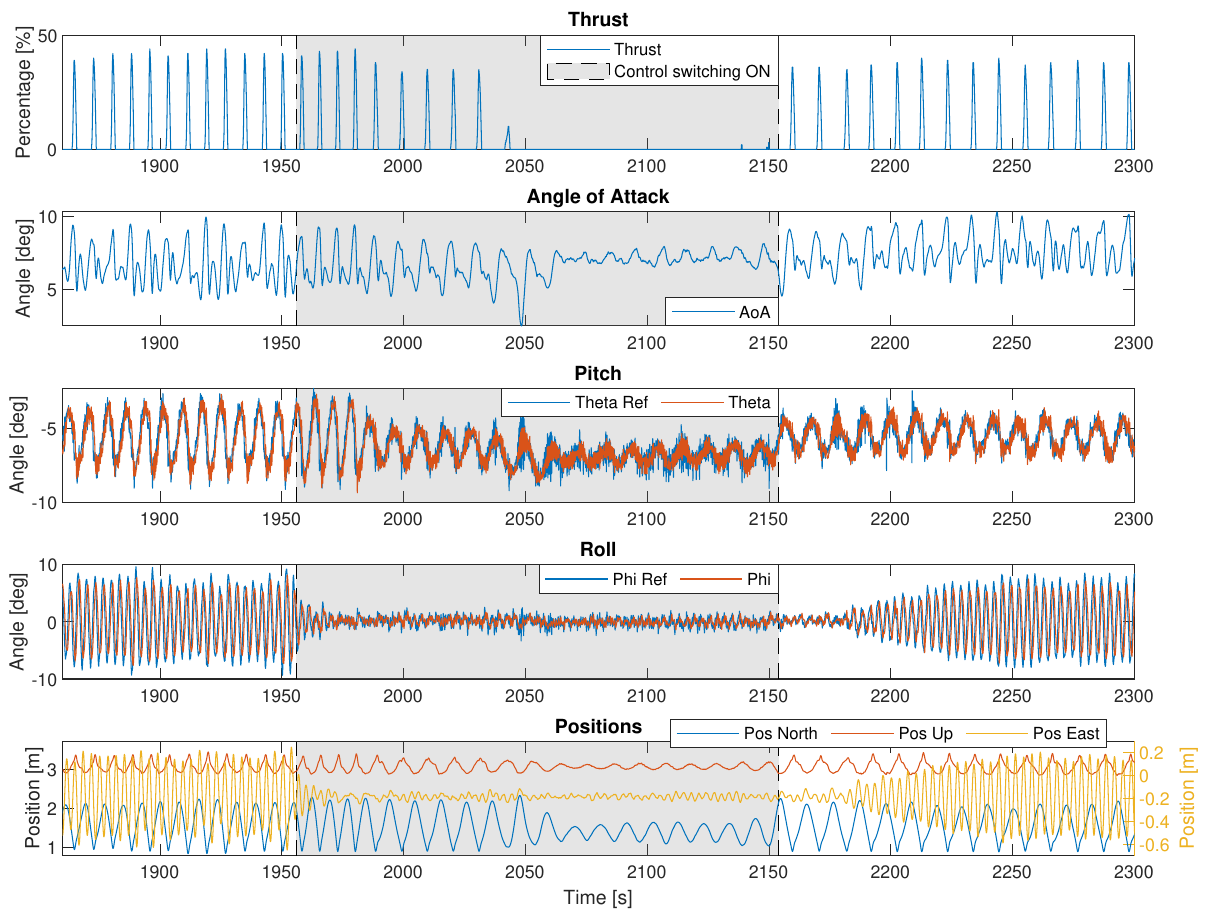}
\caption{
%fr0054 eclipson soar controller switching. 
Thrust, AoA, attitude, and positions during the flight in the wind tunnel for the \textbf{\textit{Eclipson C}} plane. The gray area indicates when the control switching method was enabled. The nominal wind speed was set to 10.5 $m/s$, and the measured wind speed was 9.2 $m/s$. 
}
    \label{f:plot_fr0054}
\end{figure*}

We began the flight using the BASE controller and positioned the MAV at the reference location. Then, we switched to the SAOS and subsequently back to the BASE controller, keeping the reference position fixed throughout. This allowed for a direct comparison of control behaviors under identical conditions.
Figure \ref{f:plot_fr0054} shows the thrust usage, AoA, attitude, and positions during the flight.
At the start of the experiment, the MAV showed circling behavior, characterized by repeated position oscillations and spikes in throttle usage. It also experienced significant roll oscillations, ranging from $+5.39\deg$ to $-10.01\deg$. At 1956 seconds, the controller was switched to the SAOS. Immediately afterward, roll oscillations decreased significantly. Position errors and throttle usage also began to decrease gradually. From approximately 2050 seconds onward, throttle usage dropped to zero and remained at zero. At 2160 seconds, the controller was switched back to the BASE controller. Upon switching, throttle usage increased immediately, and circling behavior re-emerged. Despite allowing sufficient time in this state, none of the key metrics—throttle usage, roll oscillation, or position errors—showed convergence or improvement under the BASE controller.

\subsection{SEAL G1500}
\begin{figure*}[htb]
\centering
\includegraphics[width=1.0\linewidth]{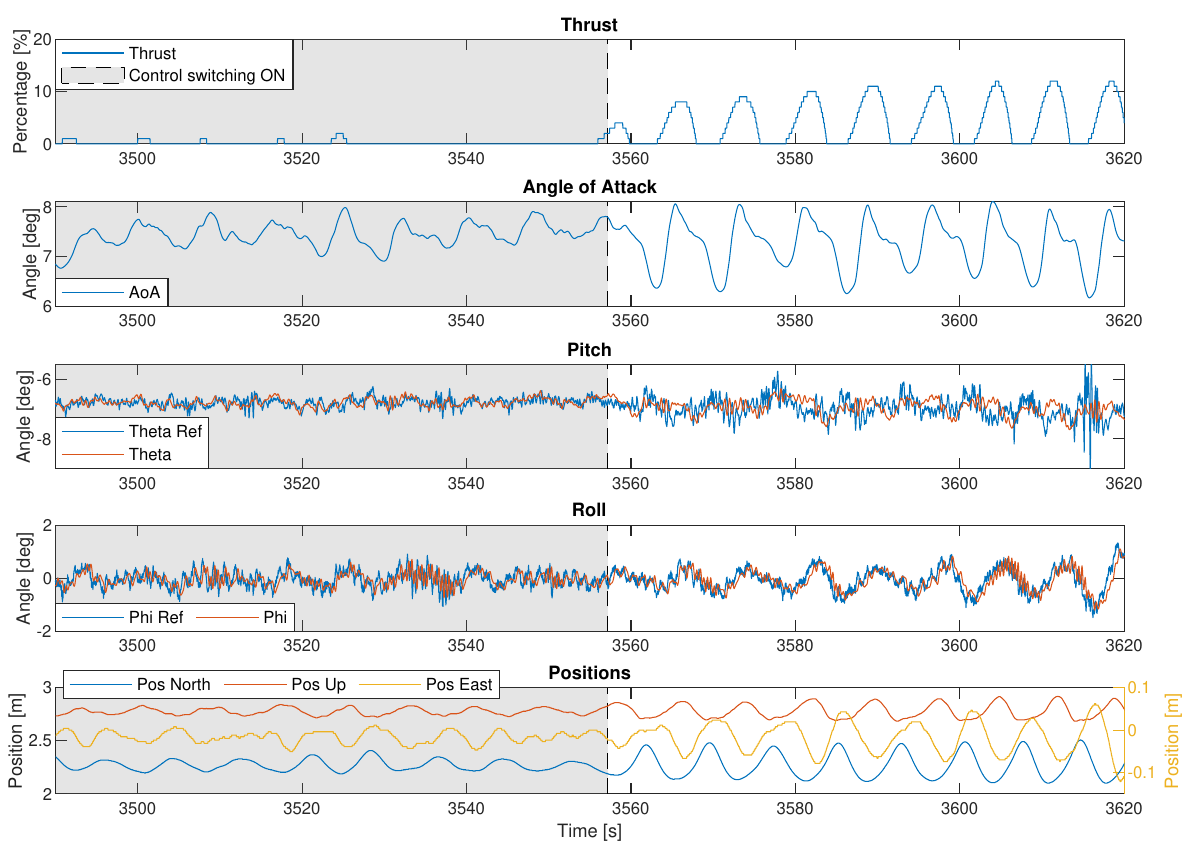}
\caption{
%fr0440-2 SEAL soar controller switching. 
Thrust, AoA, attitude, and positions during the flight in the wind tunnel for the \textbf{\textit{Seal G1500}} plane. The gray area indicates when the control switching method was enabled. The nominal wind speed was set to 12 $m/s$, and the measured wind speed was 10.7 $m/s$. 
}
    \label{f:plot_fr0440_2}
\end{figure*}

The Seal plane has a longer wingspan, a larger fuselage, and greater weight compared to the other platform. Its extended wingspan left less lateral margin to remain within the wind section. As a result, flights with the BASE controller frequently had to be terminated, as the MAV drifted outside the wind region due to pronounced roll oscillations. The lateral position instability also made it difficult to initiate flights using the BASE controller.
Figure \ref{f:plot_fr0440_2} shows the thrust usage, AoA, attitude, and positions during the flight.
Therefore, in a separate run, we began the flight using the SAOS. This allowed the MAV to achieve stable soaring flight with near-zero throttle usage. After switching to the BASE controller, the MAV began to exhibit circling behavior, with increasing oscillations and throttle usage over time. The flight was eventually terminated due to growing position errors, particularly in the lateral direction. The moment the MAV diverged and exited the wind section from this flight is shown in figure \ref{f:main_pic_seal}-(b).

\section{DISCUSSION}

The flight test results confirm that the proposed SAOS significantly improves both stability and energy efficiency during orographic soaring, compared to the BASE controller. By dynamically switching control authority based on throttle availability, the MAV was able to eliminate persistent circling behavior, reduce roll oscillations, and maintain its position with minimal energy expenditure. These findings highlight the importance of adapting control strategies in wind-hovering scenarios where the throttle is saturated and control surface authority is limited. 
While the indoor test flights and simulations did not reveal a significant advantage from angle-of-attack limiting, this stall-prevention feature is expected to be more beneficial in outdoor environments. In such settings, the MAV typically undergoes more aggressive maneuvers and experiences larger position deviations, increasing the risk of stall. Consequently, stall prevention may play a more critical role in maintaining stability during outdoor flights.
Notably, the method’s effectiveness across two MAV platforms with distinct geometries and dynamics supports its generality and potential for broader application.
%The use of angle-of-attack sensing and enhanced aerodynamic modeling further contributed to stall prevention and improved force estimation. Notably, the method’s effectiveness across two MAV platforms with distinct geometries and dynamics supports its generality and potential for broader application.

Although it was possible to perform wind-hovering with the BASE controller, it often failed to converge to the reference position and diverged over time. This issue was especially pronounced with the larger Seal platform, which struggled to maintain lateral position within the confined wind tunnel environment and frequently exited the wind section due to excessive roll oscillations. %The limited lateral space in the test setup amplified this instability, but the same risk exists in real-world applications—such as soaring near cliffs, hills, or buildings—where lateral margins are naturally constrained.

Analysis of the flight logs revealed substantial differences in roll-axis behavior between the BASE and SAOS. Divergence in position was not solely a function of horizontal or vertical control performance but was strongly influenced by lateral dynamics, especially under throttle-limited conditions. These roll oscillations are caused by coupling between the roll, pitch, and acceleration errors in the control effectiveness model. When the throttle is zero or negatively saturated $(T=0)$, the relationship between accelerations and attitude increments is given by:
\begin{equation}
    \begin{split}
        \Delta a_x = (s\psi c\phi L)\Delta\phi + (s\psi s\phi L_d)\Delta\theta \\
        \Delta a_z = (-s\phi L)\Delta\phi +(c\phi L_d)\Delta \theta 
    \end{split}
    \label{eq:roll_oscillation}
\end{equation}

Under our typical flight conditions, the yaw angle ($\psi$) remains small. In this regime, the influence of both pitch ($\Delta \theta$) and roll ($\Delta \phi$) on $\Delta a_x$ is significantly reduced, as their contributions are scaled by $sin\psi$. Meanwhile, vertical acceleration $\Delta a_z$ remains largely controllable through pitch. When the controller prioritizes vertical motion (z-axis), it primarily reduces z-axis acceleration error by allocating pitch. However, with pitch being ineffective for controlling $\Delta a_x$ at low yaw, and now committed to controlling $\Delta a_z$, the controller attempts to use roll to compensate for the remaining x-axis error. Since roll also has limited effectiveness in influencing $\Delta a_x$ under small yaw, this leads to large roll commands, and these roll inputs result in unintended roll oscillations, despite the desired motion being primarily in the x-direction.

However, by using the switching method, the controller minimizes either the x-axis or z-axis error exclusively, rather than attempting to balance both simultaneously. When the z-axis is prioritized, the controller primarily utilizes pitch to regulate vertical acceleration, without compromising by introducing roll to compensate for errors in the x-axis. Consequently, the switching method not only enhances the effectiveness of pitch control but also contributes to roll stability by reducing unnecessary roll input. Overall, with BASE, roll oscillations were frequently observed when the MAV reduced the throttle to zero near the feasible soaring area. In contrast, SAOS did not exhibit aggressive roll behavior, resulting in a more stable lateral position.

The presence of roll oscillation in the BASE controller not only degraded control performance but also increased the risk of crashing, especially in environments with limited lateral clearance. This underscores a broader point: the challenges addressed by the control switching method are not limited to the specific wind tunnel setup, but are relevant for practical orographic soaring. 

The limited lateral space in the test setup amplified the lateral instability, but the same risk exists in real-world applications—such as soaring near cliffs, hillsides, or buildings—where the lateral axis plays an equally critical role. These environments often impose natural constraints on horizontal maneuvering space, making lateral stability essential for safe and efficient flight.

\section{CONCLUSION}

Orographic soaring offers significant benefits for MAVs, enabling increased flight endurance and extended mission or surveillance capability. However, a unique challenge associated with orographic soaring is the circling flight behavior, which increases energy expenditure and raises the risk of crash or fly-away. Addressing this behavior is therefore critical.

To mitigate circling and improve control performance, we proposed a control switching method with an additional drag term and force calculations based on AoA in the control effectiveness matrix, also referred to as \textbf{\textit{SAOS}}. This approach effectively transforms the underactuated control structure into a fully actuated one by eliminating control conflicts between the horizontal and vertical axes. In addition, we modified the control effectiveness matrix in the outer loop of the INDI controller to incorporate AoA for more accurate force modeling.

We validated these methods through simulation for various control configurations, using 300 randomized initial positions to test each configuration. The statistical results demonstrate that the switching method significantly improves position accuracy and reduces throttle usage. Furthermore, the proposed method was implemented on two MAV platforms and validated in wind tunnel experiments.

The control switching method consistently led to better convergence to the soaring position, reduced circling behavior, and lower energy consumption. In contrast, deactivating the SAOS and switching back to the BASE controller resulted in immediate roll oscillations, increased circling, and, in some cases, complete divergence from the wind section.

Roll axis oscillations were frequently observed during wind tunnel flights with the BASE controller. These were caused by a coupling between pitch and roll in the control effectiveness matrix when the controller attempted to simultaneously regulate both longitudinal and vertical-axis accelerations. This behavior is not only inefficient but can also lead to incidents in confined lateral spaces. The control switching method prevents this issue by eliminating the control conflict between longitudinal and vertical error regulation.

Future work includes conducting outdoor flight tests with the control switching method, implementing gain scheduling to enhance convergence, and incorporating aerodynamic angles into the control effectiveness matrix. In this study, we assumed the direction of aerodynamic forces remained fixed, which could be improved for greater modeling accuracy.

%\section*{Acknowledgments}
%We would like to thank Simon Cajagi for helping with the wind tunnel experiment, and MAVLab members for their support. 

% {\appendix[Proof of the Zonklar Equations]
% Use $\backslash${\tt{appendix}} if you have a single appendix:
% Do not use $\backslash${\tt{section}} anymore after $\backslash${\tt{appendix}}, only $\backslash${\tt{section*}}.
% If you have multiple appendixes use $\backslash${\tt{appendices}} then use $\backslash${\tt{section}} to start each appendix.
% You must declare a $\backslash${\tt{section}} before using any $\backslash${\tt{subsection}} or using $\backslash${\tt{label}} ($\backslash${\tt{appendices}} by itself
%  starts a section numbered zero.)}


\bibliographystyle{IEEEtran}
\bibliography{root}

% \newpage

 

% \end{IEEEbiography}



\vfill

\end{document}